\documentclass[a4paper,conference,anonymous,review,screen]{IEEEtran}
\ifCLASSINFOpdf
\else
\fi
\usepackage[switch]{lineno}
\usepackage{booktabs}
\usepackage{multirow}
\usepackage{enumerate}
\usepackage{bbding}
\usepackage{epsfig}
\usepackage{graphicx}
\usepackage{amsmath}

\usepackage{array}
\usepackage{color}
\usepackage{url}
\usepackage{subfigure}
\usepackage{bm}
\usepackage{verbatim} 
\usepackage{multirow}
\usepackage{booktabs} 

\usepackage{setspace}
\makeatletter

\begin{document}
%
\title{Multi-scale 2D Representation Learning for weakly-supervised moment retrieval}

\author{\IEEEauthorblockN{Ding Li$^{1,2}$, Rui Wu$^3$, Yongqiang Tang$^1$, Zhizhong Zhang$^{1,2}$, Wensheng Zhang$^{1,2}$}
\IEEEauthorblockA{$^1$Institute of Automation, Chinese Academy of Sciences.\\
$^2$School of Artificial Intelligence, University of Chinese Academy of Sciences, Beijing, China.\\
 $^3$Horizon Robotics, China.\\
(liding2019, tangyongqiang2014, zhangzhizhong2014)@ia.ac.cn, rui.wu@horizon.ai, zhangwenshengia@hotmail.com}
}


%


\maketitle

\begin{abstract}
Video moment retrieval aims to search the moment most relevant to a given language query. However, most existing methods in this community often require temporal boundary annotations which are expensive and time-consuming to label. Hence weakly supervised methods have been put forward recently by only using coarse video-level label. Despite effectiveness, these methods usually process moment candidates independently, while ignoring a critical issue that the natural temporal dependencies between candidates in different temporal scales. To cope with this issue, we propose a Multi-scale 2D Representation Learning method for weakly supervised video moment retrieval. Specifically, we first construct a two-dimensional map for each temporal scale to capture the temporal dependencies between candidates. Two dimensions in this map indicate the start and end time points of these candidates. Then, we select top-K candidates from each scale-varied map with a learnable convolutional neural network. With a newly designed Moments Evaluation Module, we obtain the alignment scores of the selected candidates. At last, the similarity between captions and language query is served as supervision for further training the candidates' selector. Experiments on two benchmark datasets Charades-STA and ActivityNet Captions demonstrate that our approach achieves superior performance to state-of-the-art results.
\end{abstract}


%
\IEEEpeerreviewmaketitle

\section{Introduction}

Video moment retrieval can facilitate a lot of multimedia applications, \textit{e.g.} video surveillance, sport analytics and short-term video recommendation. 

Therefore, it has drawn much research interest in recent years\cite{gao2017tall, zhang2019learning,wang2019temporally,wang2019language}. This task aims to search the moment most relevant to the given text query in an untrimmed video. Taking Fig. 1 as an example, given a text query "A person is eating a sandwich. ", we want to know when this event starts and ends in the whole video. 

\begin{figure}[t]
	\centering
	\includegraphics[width=8.2cm,height=5.0cm]{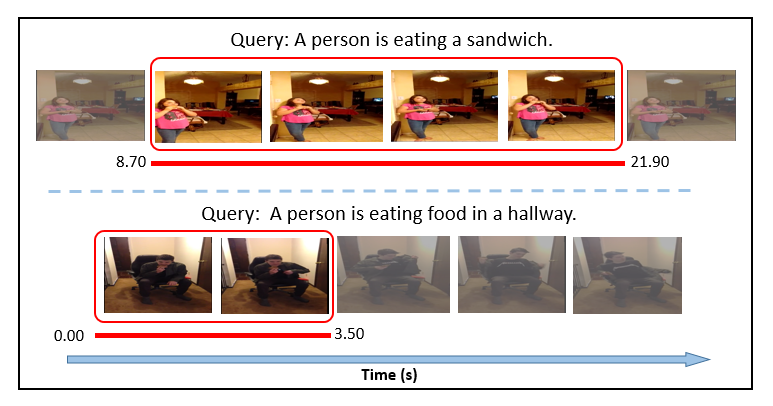}
	\caption{Video moment retrieval task: Localize the best-matched video moments in an untrimmed video for the given text query.}
	\label{fig:introduction}
\end{figure}

During the past several years, deep learning based approaches have greatly promoted the development of video moment retrieval. Most of these methods use a fully-supervised training manner, which requires accurate annotations of the start and end time points of the corresponding moments for given text queries. However, manually labelling temporal boundary of the moments is time-consuming and of high cost. Besides, the temporal boundaries of moments are usually ambiguous to define, which brings more difficulties for accurate labelling. To remedy the above issues, more
recently, intense attention \cite{mithun2019weakly,lin2019weakly-supervised,tan2020logan,gao2019wslln} is being paid for developing a weakly-supervised training mechanism, which merely requires video-level description for training data and thus leads to the significant cost saving. Although several works have made in the weakly-supervised setting, there are still two crucial issues requiring to be handled.

First, most existing weakly-supervised methods resort to projecting the text features and video features of moment candidates into some learned unified space, and then calculate the alignment score of candidates with text query, the larger score indicates the higher probability to be the result. However, these methods process each moment candidate individually, thus the relations between video moments are inevitably neglected. Second, the present weakly-supervised moment retrieval methods generally overlook the fact that the variance of temporal scale of video moments is also an important influence factor for moments localization. For example, both text queries shown in Fig.1 are devoted to describing a similar event of person eating sandwich, but the temporal lengths of the corresponding video moments varies greatly. 

For the first issue, we spot that this issue has been concerned in several works for fully-supervised temporal action detection. \cite{opazo2019proposal-free} employed self-attention mechanism and update the features by aggregating the information from other candidates with learned weights, but it brings much computational cost. \cite{zeng2019graph} constructed a candidate graph updated by graph neural network(gnn) \cite{scarselli2008graph}, in which relations between moment candidates are implicitly represented by the edges between candidate nodes. The constructed graph does not characterize temporal dependencies between nodes explicitly, \cite{zhang2019learning} then proposed the 2D-TAN method, which consists of a single-scale 2D temporal feature map to explicitly represent and capture the temporal dependencies between moment candidates and has achieved promising performance. However, the lack of temporal boundary annotations is not conductive to handle the variance of temporal scale and design loss function in training time, which makes it difficult to directly transfer the proposed framework in these fully-supervised methods into the weakly-supervised task. For the second issue, in fully-supervised setting, \textit{e.g.} 2D-TAN, such challenge is weakened by the strong supervision of massive accurate moment annotations, but we note that the single-scale 2D map applied in \cite{zhang2019learning} achieved limited success when generate more precise moment candidates, which contributes little to the weakly-supervised training.

These considerations motivate us to propose a multi-scale 2D Representation Learning model for Weakly-supervised moment retrieval. The key idea is to construct multiple 2D temporal feature maps with different temporal sampling scale, and then evaluate the alignment scores of moment candidates. The representation learning model resort to the two-stage pipeline, and consists of Multi-scale 2D Temporal Network and Moment Evaluation Module. In the Multi-scale 2D Temporal Network, we construct the Multi-scale 2D Temporal Map and perform convolution over the map to capture the temporal context. In the Moment Evaluation Module, we introduce the video caption module for each input moment candidate, and generate pseudo label for training. We evaluate our proposed method on two popular benchmark datasets for video moment retrieval, \textit{e.g.} Charades-STA and ActivityNet Captions Dataset. 

The main contribution of this paper can be concluded as follows: 
\begin{enumerate}[1.]
	\item We introduce a novel multi-scale 2D temporal network, which elaborate multi-granularity moment candidates generation and captures temporal dependencies between moment candidates. 
	\item We propose a moment evaluation module with reconstruction-guided binary cross-entropy loss (RG-BCE loss), which facilitates the weakly-supervised training.
	\item Experiment results on the two benchmark datasets (Charades-STA and ActivityNet Captions) verify the effectiveness of our proposed method. 
\end{enumerate}

The rest of this paper is organized as follows. In Section \uppercase\expandafter{\romannumeral2}, we briefly review some related works, followed by the introduction of proposed multi-scale representation learning method in Section \uppercase\expandafter{\romannumeral3}. Experimental results and discussions are showed in Section \uppercase\expandafter{\romannumeral4}. Finally, we conclude this paper in Section \uppercase\expandafter{\romannumeral5}.

\begin{figure*}[]
	\centering
	\includegraphics[width=17.8cm,height=8.8cm]{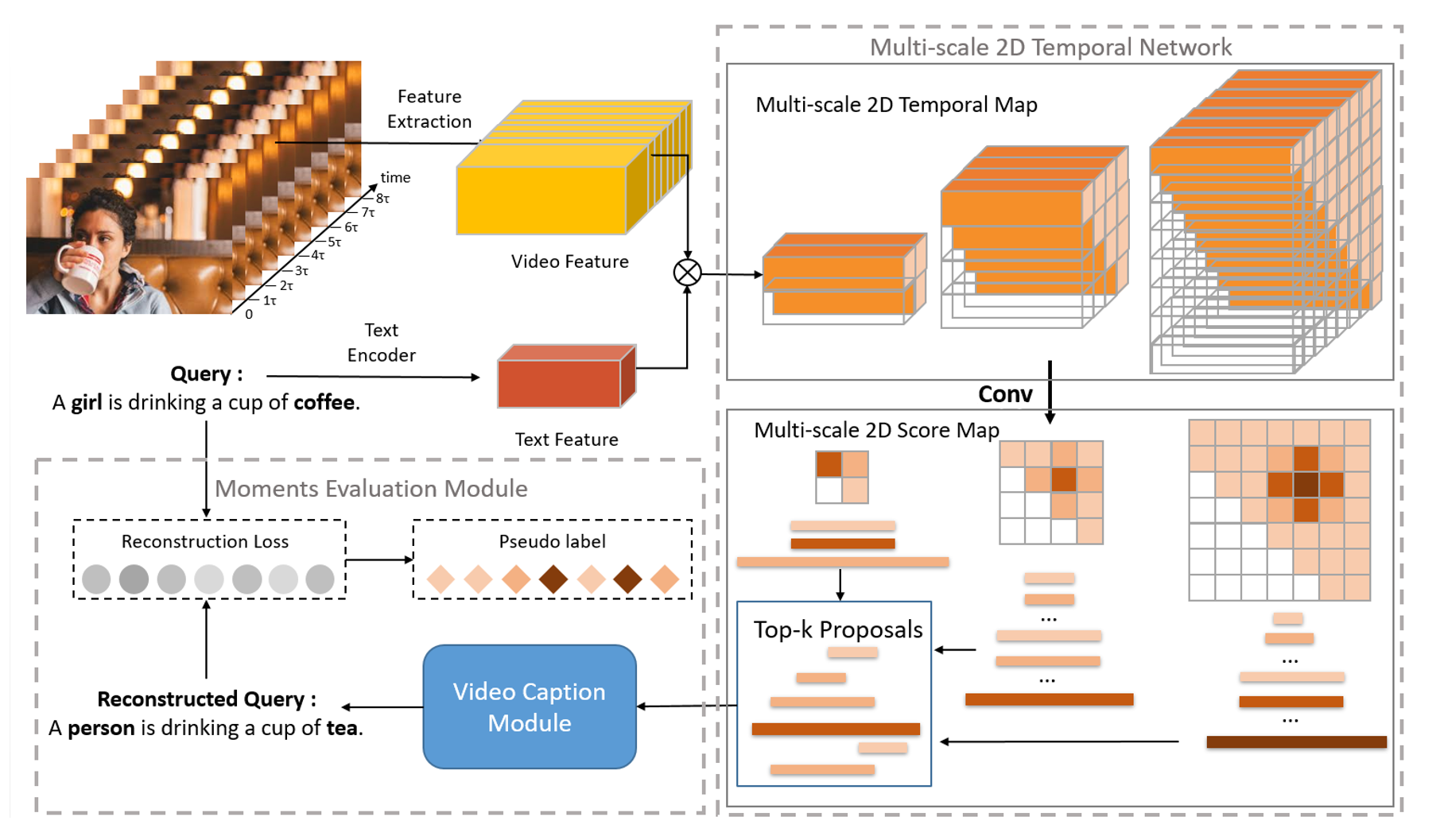}
	\caption{The Framework of our Multi-scale  Representation Learning for weakly-supervised moment retrieval. Taking a video-sentence pair as input, we extract the basic video and text representations by Feature Extractor and Text Encoder. After that, we construct the Multi-scale 2D Temporal Network which consists of 2D feature map and 2D score map. Then, we reconstruct the text query based on the top-k moment candidates in the map, and generate the pseudo labels for training.}
	\label{fig:framework}
\end{figure*}

\section{Related Works}
In this section, we will mainly focus on the related works of temporal action detection and recent related advances in video moment retrieval via text queries.

Temporal action detection aims at localizing boundaries and classifying category of action instances in untrimmed videos. The two-stage method first generates action instances with temporal boundaries and followed by classifier. These works mainly focus on generating proposals with precise boundaries. \cite{lin2018bsn} adopted three activeness curves to locate flexible proposal boundaries, \cite{zeng2019graph} used graph network to extract features between different proposals,  \cite{lin2020fast}used two feature maps separately for completeness regression and temporal boundary classification. By contract, the one-stage method integrates location and classification into a single step and hence achieves higher efficiency.

Besides, the weakly-supervised temporal action localization only uses video-level action category as label when detecting the temporal boundaries. Autoloc \cite{shou2018autoloc} regressed the confidence scores and then generates more accurate proposals. BaS-Net \cite{lee2019background} proposed an asymmetrical two-branch weight-sharing architecture to handle the background. However, the temporal action are limited to the pre-defined simple action category, which is not flexible to some video understanding applications.

To overcome aforementioned limitation of temporal action detection, Gao \cite{gao2017tall} and Hendricks \cite{anne2017localizing} introduced the video moment retrieval via text queries. \cite{gao2017tall} proposed to jointly model video clips and text queries using multi-modal operations, then alignment scores and location offsets were predicted based on the multi-model representation. \cite{anne2017localizing} proposed to embed both modalities into a common space and minimize the squared distances. \cite{xu2019multilevel} followed a two-stage pipeline to retrieve video clips. They first generated query-specific proposals from the videos, then utilized caption reconstruction. In \cite{chen2019semantic}, a visual concept based approach was proposed to generate proposals, followed by proposal evaluation and refinement. \cite{wang2019language} explored reinforcement learning to find the corresponding segments. \cite{zhang2019learning} introduced the 2d temporal feature map to represent the moment candidates with temporal relations, and achieved better performance.

Inspired by the success of the weakly-supervised temporal action detection, a small number of works are proposed to retrieve best-matching video moment without annotations of temporal boundaries. \cite{duan2018weakly} decomposed the problem of weakly-supervised dense event captioning in videos into a cycle of dual problems: caption generation and moment retrieval, and explores the one-to-one correspondence between the temporal segment and event caption. \cite{mithun2019weakly} proposed a weakly-supervised joint visual-semantic embedding framework for moment retrieval, and utilizes the latent alignment for localization during inference. \cite{tan2020logan} exploited a multi-level co-attention mechanism which comprises of a Frame-By-Word interaction module as well as a novel Word-Conditioned Visual Graph (WCVG), and incorporate the positional embedding in the temporal sequence. \cite{gao2019wslln} designed an alignment branch and a detection branch, and merge the moment-text matching score for the evaluation. \cite{lin2019weakly-supervised} constructed a novel semantic completion network for moment candidates evaluation, and exploited the alignment relationship. However, these methods processed the moment candidates individually, which neglects the temporal context information.

\section{Approach}

In this section, we first introduce the problem definition of this task, and then present the pipeline for the Multi-scale 2D Representation Learning, including Multi-scale 2D Temporal Network and Moments Evaluation Module. In the proposed pipeline, we utilize a multi-scale 2D temporal map to represent the video segments, and employ 2D Temporal Network on the constructed maps. Then, top-k moment candidates are selected from the set of maps and taken as input of Moments Evaluation Module. Eventually, we can get the final video moments according to the evaluation score.

\subsection{Problem Definition}

As mentioned before, the goal of this paper is to retrieve video moments of interest in a weakly-supervised setting. Given a video denoted as $V = \left\{ {{v_i}} \right\}_{i = 1}^{{N_v}}$ and a sentence $T = \left\{ {{t_i}} \right\}_{i = 1}^L$ as text query, we aim to automatically retrieve the most relevant video segment according to the query. $N_v$ is the number of the frames of the video, and $L$ is the length of the sentence. Specifically, we can get the best-matched moment $M = \left\{ {{\tau _s},{\tau _e}} \right\}$, where ${\tau _s},{\tau _e}$ are the indices of start and end frame respectively. Note that there is no need to have access to the temporal boundary annotations of video moments in the training time.

\subsection{Basic Video and Text Representation}
This section introduces the basic feature representation of the input text query and untrimmed video.

\textbf{Video Representation. }As for the given untrimmed video $V = \left\{ {{v_i}} \right\}_{i = 1}^{{N_v}}$, we first split the whole video into several video clips, then each video clip would be used as the input of a pre-trained 3D CNN model. In the procedure of video split, we utilize the multiple fixed intervals to sample frames from the original video, which facilitates the construction of the multi-scale 2D map. Following the setting of \cite{zhang2019learning}, spatio-temporal feature are extracted by the pre-trained 3D CNN model, and then passed through a fully-connected layer with ${d^v}$ output channels. 

\textbf{Text Representation. }The text encoder includes the word embedding and LSTM network \cite{gers1999learning}. We use GloVe word2vec model to extract the word embedding of each word in the input sentence. For each word ${t_i}$ in the input sentence, the respective embedding vector are generated as ${w_i} \in {R^{{d^T}}}$, ${d^T}$ is the length of the vector. The embedding vector $\left\{ {{w_i}} \right\}_{i = 1}^L$ are then fed into the three-layer bidirectional LSTM network, and we utilize the last hidden state as the text representation of the sentence. The final text feature are extracted as ${f^T} \in {R^{{L \times d^T}}}$, which encodes the input text query.

\subsection{Multi-scale 2D Temporal Network}

The Multi-scale 2D Temporal Network takes an the basic video and text representation as input, and outputs ${N_s} \times K$ segment proposals and corresponding alignment scores respectively. ${N_s}$ is the number of scales, and $K$ is the number of selected segment proposals in each scale. 

To get the segment proposals more precisely, we perform multi-scale temporal sampling on the untrimmed video $V = \left\{ {{v_i}} \right\}_{i = 1}^{{N_v}}$. Specifically, we first segment it into small video clips. Each video clip consists
of $T$ frames. Then, we repeatedly sample the video clips with $N_s$ intervals. After j-th sampling, we get $N_j$ video clips, and the original video would be converted to $V = \left\{ {S_i^j} \right\}_{i = 1,j = 1}^{{N_j},{N_s}}$, where $S_i^j$ is the sampled clips. Further more, we extract the deep 3D CNN feature of each clip as mentioned before, denoted as $\left\{ {{f^S}} \right\}_{i = 1,j = 1}^{{N_j},Ns}$. To get a more compact representation, we pass the extracted feature through a fully-connected layer with $d^v$ output channels. For the i-th video segment sampled in j-th scale, the final 3D feature is ${f^S} \in {R^{{d^v}}}$, where ${d^v}$ is the feature dimension. 

The video segments obtained by multi-scale temporal sampling are set as the input of the multi-scale 2D temporal feature map construction, and each grid in the map represents a moment candidate with start and end indexes along the axis. The moment feature of each grid in the 2D map are extracted on the basis of the 3D segment feature ${f^S} \in {R^{{d^v}}}$. In the extraction process, we follows the temporal pooling design. For each moment candidate, we perform max-pooling operation on the corresponding segments with the reference of the start and end indexes in the 2D map. For the moment in the x-th row and y-th column of the map, we can obtain the feature of this moment candidate: 
$$F_{x,y}^j = \left\{ {\begin{array}{*{20}{l}}
	{\max {\mathop{\rm pool}\nolimits} \left( {f_x^S,f_{x + 1}^S, \cdots ,f_y^S} \right),}&{if 0 < x \le y < {N_j}}\\
	{{0^S},}&{else}
	\end{array}} \right.$$
where this moment starts from time stamp $x$ and ends in time stamp $y$. When $0 < x \le y < {N_j}$, the value of the moment feature is non-zero. Thus, the 2D map of the $j$-th scale is constructed, and the corresponding moment features in the map are extracted by aggregating the video segment features.

We denote the the 2D temporal feature map of the j-th scale as ${F^j} \in {R^{{N_j} \times {N_j} \times {d^v}}}$. ${N_j}$ is the number of sampled segments in the j-th temporal scale, and also represents the start and end indexes in the j-th 2D map. Different from the single-scale 2D temporal feature map, we collect all the 2D map constructed in multiple scales, denoted as ${F^M} = \left\{ {{F^j}} \right\}_{j = 1}^{{N_s}}$. 

To select the proper candidate moments, we need to sample the possible moments based on the 2D temporal feature map. As introduced in \cite{zhang2019learning}, one simple way is to enumerate all the possible consecutive video clips as candidates. While the other way is to sparsely sample the moments, this could efficiently remove the redundant moment candidates, and save the computational cost simultaneously. So we choose the latter sampling strategy, and get the selected moment candidates as proposals.

The 2D Temporal Network mainly consists of the cross-modal fusion and the convolution network over the multi-scale 2D temporal feature map, and update the cross-modal feature by capturing the temporal dependencies on the 2D map.

After the construction of the multi-scale 2D temporal feature map which represent the video moment candidates, the next step is the cross-modal fusion based on the text query features. In order to align the text query with moment candidates in multi-scale 2D map, we first duplicate the extracted text embedding feature ${f^T}$ $N_s$ times, denoted as ${F^T} = {\left\{ {{f^T}, \cdots ,{f^T}} \right\}_{{N_s}}}$. Then, the text features and video moment features in the map are projected into a common feature space by a fully-connected network. And eventually the cross-modal feature map are fused by the Hadamard product and ${l_2}$ normalization. Mathematically, the generation of the cross-modal feature can be formulated as follows:
$${F_{cro}} = {\left\| {\left( {{W^T} \cdot {F^T} \cdot {1^T}} \right) \odot \left( {{W^M} \cdot {F^M}} \right)} \right\|_F},$$ 
where ${W^T}$ and ${W^M}$ represents the parameters of the fully connected layers, which could be learnt in training. ${1^T}$ is the transpose of an all-ones vector, $\odot$ is Hadamard product, and ${\left\|  \right\|_F}$ denotes Frobenius normalization.

In order to capture the temporal dependencies between moment candidates in the multi-scale 2D feature map, we construct the multi-layer convolution network on the 2D cross-modal feature map $F_{cro}$. Through the convolution network, the context information is aggregated from the adjacent moment candidates, and the enlarged receptive field makes it easy to leverage the long-term relations in the video. 

\subsection{Moments Evaluation Module}

We generate the moment candidate alignment scores by the Multi-scale 2D Temporal Network, and select the best-matching moments by this score. However, there is no full annotations of temporal boundaries served as the supervision of the moments evaluation when training in a weakly-supervised manner. Thus, we propose a Moments Evaluation Module and generate the pseudo labels for the moment candidate in the 2D multi-scale score map. In this module, we first select the top-k moment candidates with higher alignment scores from the 2D feature map in each temporal scale, and then perform moments caption based on the selected moments. According to the similarity between the reconstructed text query generated by the moments caption model and the original text query in annotations, we could obtain the pseudo labels as supervision for training. 

\textbf{Moments Caption.} Observing that the only annotation of this task is the text query, we design the caption module and make the whole model trainable. The framework of Moments Caption has been shown in Fig. 3. 

\begin{figure}[t]
	\centering
	\includegraphics[width=8.3cm,height=4.4cm]{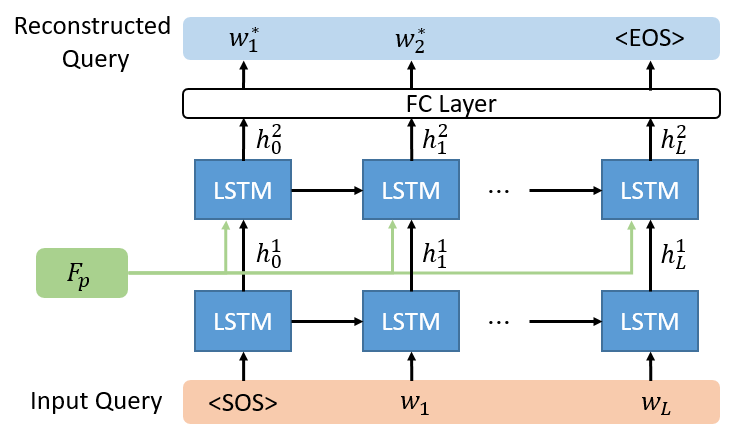}
	\caption{The Caption Module for reconstruction of the text query. This module consists of a two-layer LSTM network and a fully-connected layer, and generate the caption of the moment candidates based on their cross-modal feature and text query feature.}
	\label{fig:caption_module}
\end{figure}

The text embedding vectors ${w_i} \in {R^{{d^T}}}$ are passed through the first LSTM layer, and the hidden state $\left\{ {h_0^1,h_1^1, \cdots ,h_L^1} \right\}$ are fused with the cross-modal feature of the selected top-k moment candidates $F_cro$. Then the fused sequence feature are fed into the second LSTM layer \cite{gers1999learning}, and get the hidden state $\left\{ {h_0^2,h_1^2, \cdots ,h_L^2} \right\}$. Finally, we utilize the fully-connected layer and obtain the embedding vector of reconstructed query ${w^*_i} \in {R^{{d^T}}}$. 

\textbf{Moments Evaluation.} 
As for moments evaluation, the output of the convolution network in the 2D temporal network is passed through a fully-connected network with sigmoid activation function, and obtain the multi-scale 2D score maps, denoted as $P = {\mathop{\rm sigmoid}\nolimits} \left( {{W^F} \cdot {F_{cro}}} \right)$. Each grid in the score map represents the alignment score between the moment candidate and the given text query. Owing to the lack of temporal boundary annotations, we are not able to compute the tIoU between moment candidates and truly-matched moments. Thus, we generate the pseudo labels for further training of the evaluation module, details are illustrated in the loss functions section.

\subsection{Loss functions}

In this section, we mainly introduce the loss functions for training the framework. The Loss functions in the proposed model consist of a reconstruction loss and a cross-entropy loss. The former is calculated between text query and reconstructed query, and the latter is calculated after the moment candidates evaluation on the 2D multi-scale score map. 

Given a video-sentence pair, the reconstruction loss maximizes the normalized log likelihood of the words in the reconstructed query, denoted by:
$$ {L_{rec}} =  \frac{-1}{{{N_s}KL}}\sum\limits_{k = 1}^{{N_s}K} {\sum\limits_{l = 1}^L {\log P\left( {w_l^*\left| {F_{cro}^k,h_{l - 1}^2,{w_1}, \cdots ,{w_{l - 1}}} \right.} \right)} } $$

The reconstruction loss reflects the similarity of the input query and the reconstructed query, the learnt model tends to select moments candidates with lower reconstruction loss. For the k-th moment candidate in the j-th scale 2D temporal feature map, the reconstruction loss of the moment is denoted as $l_k^j$, and the pseudo label $y_k^j$ is computed as:
$$l_k^j = \frac{{\sum\limits_{l = 1}^L {\log \left( {w_l^*\left| {F_{cro}^k,h_{l - 1}^2,{w_1},{w_2}, \cdots ,{w_{l - 1}}} \right.} \right)} }}{{\sum\limits_{k = 1}^K {\sum\limits_{l = 1}^L {\log \left( {w_l^*\left| {F_{cro}^k,h_{l - 1}^2,{w_1},{w_2}, \cdots ,{w_{l - 1}}} \right.} \right)} } }}$$

$$y_k^j = \left\{ {\begin{array}{*{20}{l}}
	{0,}&{l_k^j \ge {l_{\max }}}\\
	{1 - l_k^j,}&{{l_{\min }} \le l_k^j < {l_{\max }}}\\
	{1,}&{l_k^j < {l_{\min }}}
	\end{array}} \right.$$

After generating the pseudo labels $y_k^j$, we adopt them as supervision of the candidates and design a reconstruction-guided binary cross-entropy loss (RG-BCE loss):
$${L_{rg - bce}} = \frac{1}{{{N_s}K}}\sum\limits_{j = 1}^{{N_s}} {\sum\limits_{k = 1}^K {y_k^j\log p_k^j + \left( {1 - y_k^j} \right)\log \left( {1 - p_k^j} \right)} } $$

In total, the final multi-task loss could be formulated as:
$$L = {L_{rg - bce}}{\rm{ + }}\lambda {L_{rec}}$$
where $\lambda$ is the hyper-parameter for balancing the reconstruction loss and binary cross-entropy loss.

\section{Experiments}
We conduct experiments on two benchmark datasets: Charades-STA and ActivityNet Captions, and evaluate the effectiveness of our multi-scale 2D representation learning for weakly-supervised video moment retrieval. We first introduce the datasets, evaluation metric and implementation details, and then report the experiment results and analysis. Finally, we discuss the impact of the parameter setting in the proposed model.

\subsection{Datasets and Evaluation Metric}

\textbf{Charades-STA.} The Charades dataset \cite{sigurdsson2016hollywood} is originally proposed in 2016. It contains 9848 videos of daily indoors activities. It is originally designed for action recognition and localization. Gao et al. extend the temporal annotation, labeling the start and end time of moments of the original video dataset with language descriptions and name it as Charades-STA. Charades-STA contains 12408 moment-sentence pairs in training set and 3720 pairs in testing set.

\textbf{ActivityNet Captions.} It consists of 19209 videos, whose content are diverse and open. It is originally designed for video captioning task, and recently introduced into the task of moment localization with natural language, since these two tasks are reversible. Following the experimental setting in \cite{zhang2019learning}, we use val-1 as validation set and val-2 as testing set, which have 37417, 17505, and 17031 moment-sentence pairs for training, validation, and testing. Currently, this is the largest dataset in this task.

\textbf{Evaluation Metric.} We use the evaluation criteria following prior works in literature \cite{mithun2019weakly,lin2019weakly-supervised,tan2020logan,gao2019wslln}. We measure rank-based performance R@K (Recall at K) which calculates the percentage of test samples for which the correct result is found in the top-K retrievals to the query sample. We follow \cite{gao2017tall} for evaluating Charades-STA and ActivityNet Captions dataset, and report results for R@1, R@5 in the condition of IoU=0.3 and IoU=0.5. 

\subsection{Implementation Details}
We utilize a three-layer LSTM for extracting the basic text features, and the feature dimension $d^v$ and $d^T$ is 512. We split the whole video into small non-overlapping video clips, and use pre-extracted C3D feature \cite{tran2015learning, jia2014caffe,karpathy2014large}for both Charades-STA and ActivityNet Captions datasets. The number of frames in one clip in Charades-STA is 4, and that in ActivityNet Captions is set to 16. The multi-layer convolution network is 8-layer with kernel size of 5. The dimension of hidden states in moment caption module is 1024, and the dimension of the Glove embedding \cite{pennington2014glove} is 300. The thresholds ${l_{\max }}$ and ${l_{\min }}$ in RG-BCE loss are respectively set to 0.7 and 0.1, and we choose the top-10 moment candidates for moments caption. We use Adam \cite{kingma2014adam} with learning rate of $1 \times {10^{{\rm{ - }}4}}$, and the batch size 128 for optimization. Non maximum suppression (NMS) \cite{neubeck2006efficient} with a threshold of 0.5 is applied during the inference.

\subsection{Quantitative Results and Analysis}
In this section, we report the quantitative experiment results and analysis on the two datasets.

\textbf{Charades-STA Dataset.} The experiment results on the Charades-STA Dataset are shown in Table 1, and we use the evaluation metric "R@n, IoU=m", where n is $\left\{ {1,5} \right\}$, and m is $\left\{ {0.5,0.7} \right\}$. 

As shown in Table 1, when comparing with other weakly-supervised approaches, our proposed method outperforms the TGA model significantly and achieves better R@1 performance compared with SCN, and the results confirm the effectiveness of context information between moment candidates in the multi-scale 2D representation learning. Through the multi-scale 2D temporal feature map, fine-grained candidates are generated and the context information between candidates is encoded by the temporal network. Although the performance of LoGAN is slightly better than ours, but it constructed a Frame-By-Word interaction and get fine-grained moments representation by co-attention with higher computational cost. 

Moreover, our proposed weakly-supervised model outperform the visual-semantic embedding approaches VSA-RNN and VSA-STV by a large margin, and also perform better than some of the fully-supervised method, which indicates our weakly-supervised model could effectively improve the performance without the annotation of the temporal boundaries. Even when comparing with the state-of-the-art fully-supervised method 2D-TAN, the margin of the prediction performance is not so large. Especially, the gap between fully-supervised 2D-TAN and our multi-scale 2D representation learning is not so large, which verifies the rationality of the designed moments evaluation module and RG-BCE loss in our approach. 

\begin{table}[]
	\setlength{\abovecaptionskip}{0.cm}
	\caption{Performance comparison results on Charades-STA Dataset. }	
	\label{table1}
	\centering
	\begin{tabular}{@{}cccccc@{}}
		\toprule
		\multirow{2}{*}{Method} & \multirow{2}{*}{Training} & \multicolumn{2}{c}{IoU0.5}                                              & \multicolumn{2}{c}{IoU0.7}                            \\ \cmidrule(l){3-6} 
		&                              & R@1                                & R@5                                & R@1                       & R@5                       \\ \midrule
		Random                  & -                            & 8.61                               & 37.57                              & 3.39                      & 14.98                     \\
		VSA-RNN                 & Full                         & 10.50                              & 48.43                              & 4.32                      & 20.21                     \\
		VSA-STV                 & Full                         & 16.91                              & 53.89                              & 5.81                      & 23.58                     \\
		CTRL \cite{gao2017tall}  & Full                         & 23.63                              & 58.92                              & 8.89                      & 29.52                     \\
		2D-TAN \cite{zhang2019learning}                 & Full                         & 39.70                              & 80.32                              & 23.31                     & 51.26                     \\ \midrule
		TGA \cite{mithun2019weakly}                    & Weak                         & 19.94                              & 65.52                              & 8.84                      & 33.51                     \\
		LoGAN \cite{tan2020logan}	                & Weak                         & \multicolumn{1}{l}{\textbf{34.68}} & \multicolumn{1}{l}{\textbf{74.30}} & \multicolumn{1}{l}{14.54} & \multicolumn{1}{l}{\textbf{39.11}} \\
		SCN \cite{lin2019weakly-supervised}                    & Weak                         & 23.58                              & 71.80                              & 9.97                      & 38.87            \\
		Ours                    & Weak                         & \textbf{30.38}                     & \textbf{69.60}                     & \textbf{17.31}            & \textbf{34.92}            \\ \bottomrule
	\end{tabular}
	
\end{table}

\begin{table}[h]
	\setlength{\abovecaptionskip}{0.cm}	
	\caption{Performance comparison results on ActivityNet Captions Dataset. }
	\label{table2}
	\centering
	\begin{tabular}{@{}cccccc@{}}
		\toprule
		\multirow{2}{*}{Method} & \multirow{2}{*}{Training} & \multicolumn{2}{c}{IoU0.3}      & \multicolumn{2}{c}{IoU0.5}      \\ \cmidrule(l){3-6} 
		&                              & R@1            & R@5            & R@1            & R@5            \\ \midrule
		Random                  & -                            & 18.64          & 52.78          & 7.63           & 29.49          \\
		VSA-RNN                 & Full                         & 39.28          & 70.84          & 23.43          & 55.52          \\
		VSA-STV                 & Full                         & 41.71          & 71.05          & 24.01          & 56.62          \\
		CTRL \cite{gao2017tall}                   & Full                         & 47.43          & 75.32          & 29.01          & 59.17          \\
		2D-TAN \cite{zhang2019learning}                  & Full                         & 59.45          & 85.53          & 44.51          & 77.13          \\ \midrule
		WS-DEC \cite{duan2018weakly}                 & Weak                         & 41.98          & -              & 23.34          & -              \\
		WSLLN \cite{gao2019wslln}                   & Weak                         & 42.80           & \textbf{-}     & 22.70           & -              \\
		SCN \cite{lin2019weakly-supervised}                     & Weak                         & 47.23          & 71.45          & 29.22          & 55.69          \\
		Ours                    & Weak                         & \textbf{49.79} & \textbf{72.57} & \textbf{29.68} & \textbf{58.66} \\ \bottomrule
	\end{tabular}
	\vspace{-0.6em}
	
\end{table}

\textbf{ActivityNet Captions Dataset.} The results in Table 2 show the performance comparison with other methods on the ActivityNet Captions Dataset, and we use the evaluation metric "R@n, IoU=m", where n is $\left\{ {1,5} \right\}$, and m is $\left\{ {0.3,0.5} \right\}$.

Similar to results on Charades-STA, compared with the weakly-supervised methods WS-DEC, WSLLN and SCN, our proposed approach has achieved better performance, and even outperform some of the fully-supervised methods. The WS-DEC method designed a iterative process of moments retrieval and caption, which leads to complicated optimization. Compared with WS-DEC, our proposed method has employed the top-k selection on the multi-scale 2D temporal feature map, and has avoided the redundant iteration. 

\begin{figure*}[ht]
	\centering
	\includegraphics[width=16.8cm,height=8.3cm]{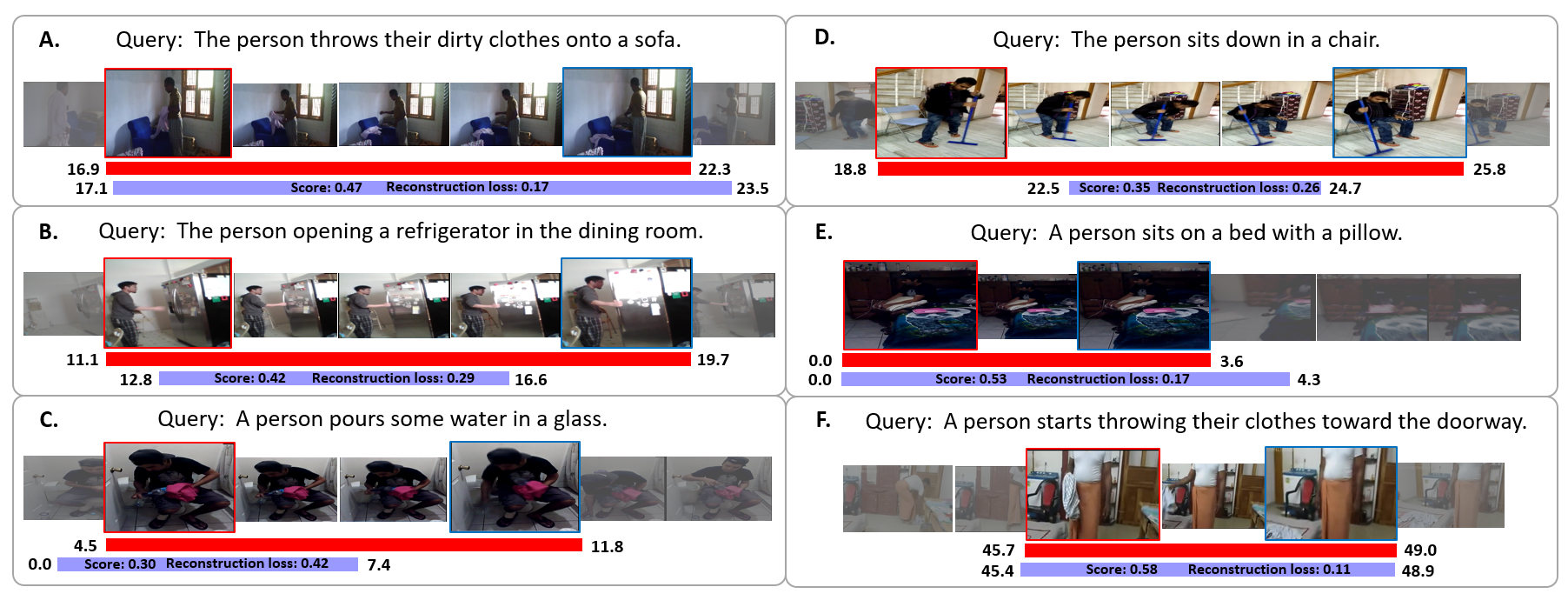}
	\caption{Qualitative results on Charades-STA datasset. The red line represents the ground truth, and the blue line is the prediction of our method. And we also demonstrate the alignment score and the corresponding reconstruction loss of the predicted video moments. }
	\label{fig:framework}
\end{figure*}

\subsection{Discussion}
In this section, we mainly discuss the impact of the selection of temporal scales and the impact of the loss weight, some of the experiment results are listed as follows.

\textbf{Impact of Multiple Temporal Scales.} To evaluate the impact of temporal scales of the 2D temporal feature map, we conduct a set of experiments outlined in Table \ref{table3}. When using our proposed multi-scale 2D temporal feature maps ($N_s=3$), the experiment results are better than that of single-scale 2D temporal feature map ($N_s=1$, $N_j=64$), which indicating the effectiveness of the multi-scale 2D feature maps. When setting $N_j=64,24,4$, the performance is boosted and better than results of single-scale map by a large margin. The smallest scale in $N_j$ is ranges in [4, 6, 8], and we get better experiment results when set it as 4, because the smaller value makes the multi-scale temporal maps able to cover more precise moments with longer temporal length.

\begin{table}[h]
	\setlength{\abovecaptionskip}{0.cm}
	\caption{Experiment results with multiple temporal scales (T-Scale). }
	\label{table3}
	\begin{tabular}{cccccc}
		\hline
		\multirow{2}{*}{T-Scale} & \multirow{2}{*}{Multi-scale} & \multicolumn{2}{c}{IoU0.3}      & \multicolumn{2}{c}{IoU0.5}      \\ \cline{3-6} 
		&                              & R@1            & R@5            & R@1            & R@5            \\ \hline
		64                               & \XSolid                      & 44.25          & 63.66          & 27.07          & 51.79          \\
		64-24-8                          & \Checkmark                   & 44.52          & 63.70          & 25.00          & 52.05          \\
		64-24-6                          & \Checkmark                   & 47.99          & 66.41          & 21.09          & 44.30          \\
		64-24-4                          & \Checkmark                   & \textbf{49.79} & \textbf{72.57} & \textbf{29.68} & \textbf{58.66} \\ \hline
	\end{tabular}
	\vspace{-0.6em}
	
\end{table}

\begin{center}
	\begin{table}[h]
		\vspace{-0.4cm}
		\caption{Experiment results with different loss weight. }
		\label{table4}
		\centering
		\begin{tabular}{ccccc}
			\hline
			\multirow{2}{*}{Loss weight} & \multicolumn{2}{c}{IoU0.3}      & \multicolumn{2}{c}{IoU0.5}      \\ \cline{2-5} 
			& R@1            & R@5            & R@1            & R@5            \\ \hline
			${\lambda}$=0.5                         & 47.09          & 74.62          & 21.63          & 54.01          \\
			${\lambda}$=1.0                         & \textbf{49.79} & 72.57          & \textbf{29.68} & 58.66          \\
			${\lambda}$=2.0                         & 43.85          & \textbf{79.98} & 24.68          & \textbf{59.67} \\ \hline
		\end{tabular}
		\vspace{-1.5em}
	\end{table}
\end{center}

\textbf{Impact of Loss Weight.} As it shows, the performance of our model is relatively stable when $\lambda$ is set as 0.5 or 1.0. When $\lambda$ = 2.0, the R@1 performance drops, while R@5 performance increases. The context information from adjacent clips would benefit the moment caption, so a few video moments with low reconstruction loss do not have the highest tIoU with ground truth, and the R@1 metric drops.

\subsection{Qualitative Results}

We present some qualitative results on the Charades-STA dataset to illustrate the effectiveness of our method, several examples are shown in Fig. 4. The red line is the ground truth, and the blue line represents the moment prediction. In Fig. 4, case A, E and F are successful cases, and the predictions in cases B, C and D are relatively misaligned compared with the ground truth. 

According to the Fig. 4, the successful moments prediction have the higher alignment scores as well as the lower reconstruction losses, which indicates that our moment evaluation module with RG-BCE loss is capable to evaluate the quality of input candidates. The successful samples include moments with long time of duration (\textit{e.g.} sample A) and those with short time of duration (\textit{e.g.} sample F), which shows the capability of our proposed method in weakening the negative affects resulting from variation of temporal scales. 

Due to the ambiguity of the action moments and existing noise in the scene, our weakly-supervised method has achieved limited success when dealing with these cases. Take sample B as an example, in the untrimmed video, a person opens the refrigerator and then closes it. It is hard to deal with ambiguity of "open" and "close" and localize the best-matching moment without temporal boundary annotations. In sample C, the noise of objects visible in the scene affects the moment evaluation, which leads to the misalignment compared with ground truth.

\section{Conlusion}
In this work, we focus on the task of video moment retrieval without manually labelling the start and end time points of moments in training. We address the motivation of considering the various temporal scale of moment candidates as well as the temporal relations between them in weakly-supervised setting, and propose a multi-scale 2D representation learning method, including the multi-scale 2D temporal network and weakly-supervised moments evaluation module with RG-BCE loss. The multi-scale 2D temporal map could generate more precise moment candidates with various temporal scales, and moment-to-text reconstruction facilitate the weakly-supervised training in moments evaluation. The experiment results on the Charades-STA and ActivityNet Captions datasets demonstrated the effectiveness and superiority of our proposed approach.

\bibliographystyle{IEEEtranS}
\bibliography{Ref}

\end{document}